\documentclass[10pt,twocolumn,letterpaper]{article}

\usepackage{cvpr}
\usepackage{times}
\usepackage{epsfig}
\usepackage{graphicx}
\usepackage{amsmath}
\usepackage{amssymb}

\usepackage[breaklinks=true,bookmarks=false]{hyperref}

\cvprfinalcopy %

\usepackage{color}

\def\softmax{{\rm softmax}}
\def\bilinear{{\rm bilinear}}
\def\Loss{{\sf Loss}}
\def\concat{{\sf concat}}
\def\downsample{{\sf downsample}}
\def\upsample{{\sf upsample}}

\def\block1{{$\sf block1$}}
\def\block2{{$\sf block2$}}
\def\block3{{$\sf block3$}}
\def\block4{{$\sf block4$}}

\def\Real{{\mathbb R}}

\def\R{{\mathbb R}}

\renewcommand{\vec}[1]{\ensuremath{\pmb{#1}}}
\newcommand{\mat}[1]{\ensuremath{\mathbf{#1}}}
\newcommand{\set}[1]{\ensuremath{\mathscr{#1}}}

\makeatletter
\@tfor\next:=abcdefghijklmnopqrstuvwxyxz\do
 {\begingroup\edef\x{\endgroup
    \noexpand\@namedef{v\next}{\noexpand\vec{\next}}%
  }\x}
\@tfor\next:=ABCDEFGHIJKLMNOPQRSTUVWXYZ\do
 {\begingroup\edef\x{\endgroup
    \noexpand\@namedef{m\next}{\noexpand\mat{\next}}%
  }\x}
\@tfor\next:=ABCDEFGHIJKLMNOPQRSTUVWXYZ\do
 {\begingroup\edef\x{\endgroup
    \noexpand\@namedef{s\next}{\noexpand\set{\next}}%
  }\x}
\makeatother

\usepackage{caption}
\begin{document}
\title{Decoders Matter for Semantic Segmentation: \\
Data-Dependent Decoding Enables Flexible Feature Aggregation
}

\author{Zhi Tian$ ^1$
        ~~~~ Tong He$ ^1$
        ~~~~ Chunhua Shen$ ^{1}$\thanks{Appearing in IEEE Conf.\
        Computer Vision and Pattern Recognition, 2019.
        First two authors equally contributed to this work. C. Shen is the corresponding author: $\tt chunhua.shen@adelaide.edu.au$}
        ~~~~ Youliang Yan$ ^2 $
\\
$ ^1$The University of Adelaide, Australia ~ ~ ~ ~ ~ ~ $ ^2$Noah's Ark Lab, Huawei Technologies
}

\maketitle
\thispagestyle{empty}

\begin{abstract}

Recent semantic segmentation methods exploit encoder-decoder architectures to produce the desired pixel-wise segmentation prediction. The last layer of the decoders is typically a bilinear upsampling procedure to recover the final pixel-wise prediction. We empirically show that this oversimple and data-independent bilinear upsampling may lead to sub-optimal results.

In this work, we propose a data-dependent upsampling (DUpsampling) to replace  bilinear, which takes advantages of the redundancy in the label space of semantic segmentation and is able to recover the pixel-wise prediction from low-resolution outputs of CNNs.
The main advantage of the new upsampling layer lies in that with a relatively lower-resolution feature map such as $\frac{1}{16}$ or $\frac{1}{32}$ of the input size, we can achieve even better segmentation accuracy, significantly reducing computation complexity. This is made possible by 1) the new upsampling layer's much improved reconstruction capability; and more importantly 2) the DUpsampling based decoder's flexibility in leveraging almost arbitrary  combinations of the CNN encoders' features. Experiments demonstrate that our proposed decoder outperforms the state-of-the-art decoder, with only $\sim$20\% of computation. Finally, without any post-processing, the framework equipped with our proposed decoder achieves new state-of-the-art performance on two datasets: 88.1\% mIOU on PASCAL VOC with 30\% computation of the previously best model; and 52.5\% mIOU on PASCAL Context.

\end{abstract}

\begin{figure}[t!]
  \centering
  \includegraphics[width=\linewidth]{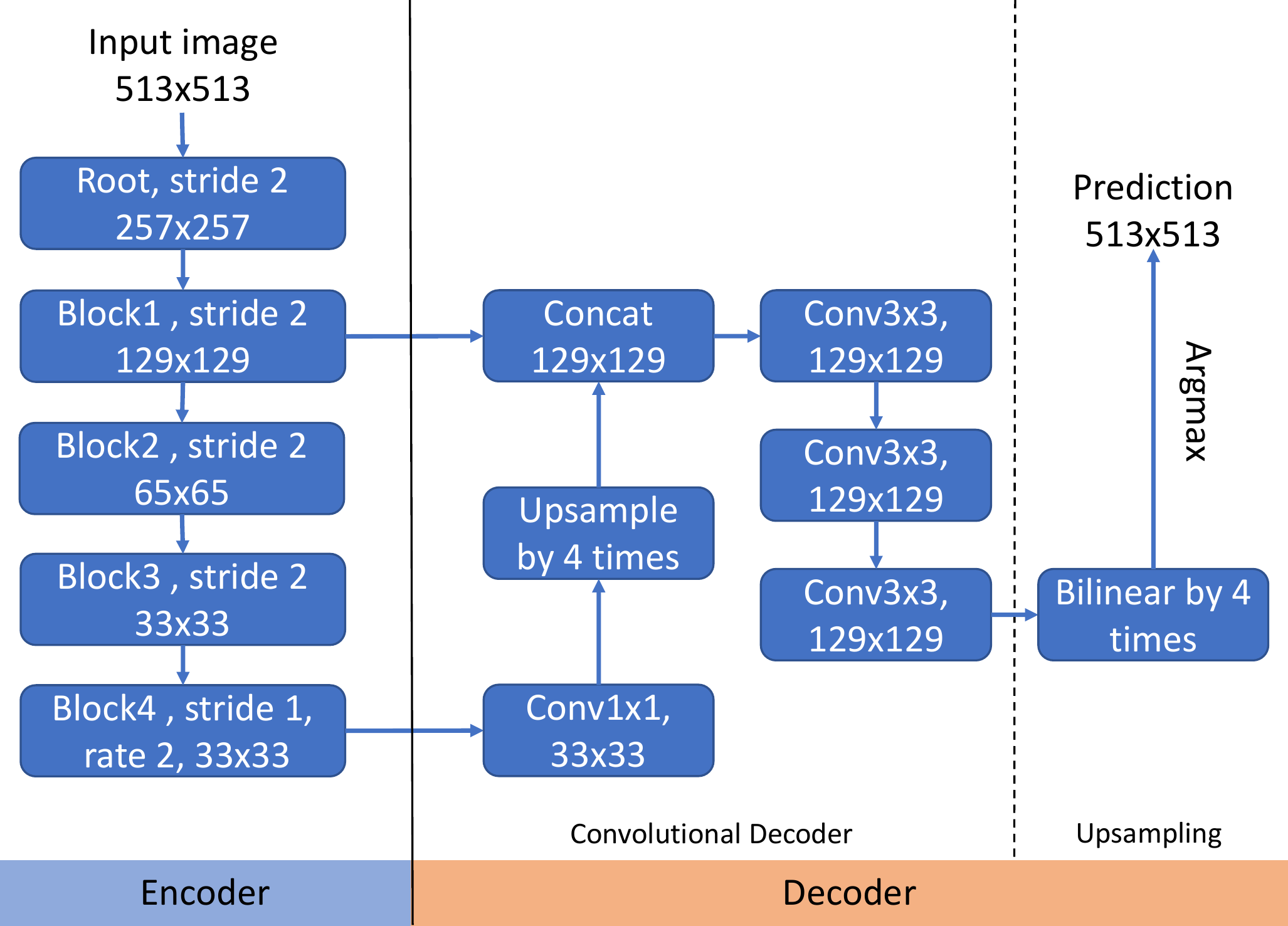}
  \caption{An example of the encoder-decoder architecture used by DeepLabv3+. Its decoder fuses low-level features of ${\sf downsample\; ratio}$ $= 4$ and upsamples high-level features before merging them. Finally, bilinear upsampling is applied to restore the full-resolution prediction. ``rate" denotes the atrous rate in atrous convolution.
  }
  \label{fig:vanilla_decoder}
\end{figure}

\section{Introduction}
\begin{figure*}[t]
  \centering
  \includegraphics[width=.9\linewidth]{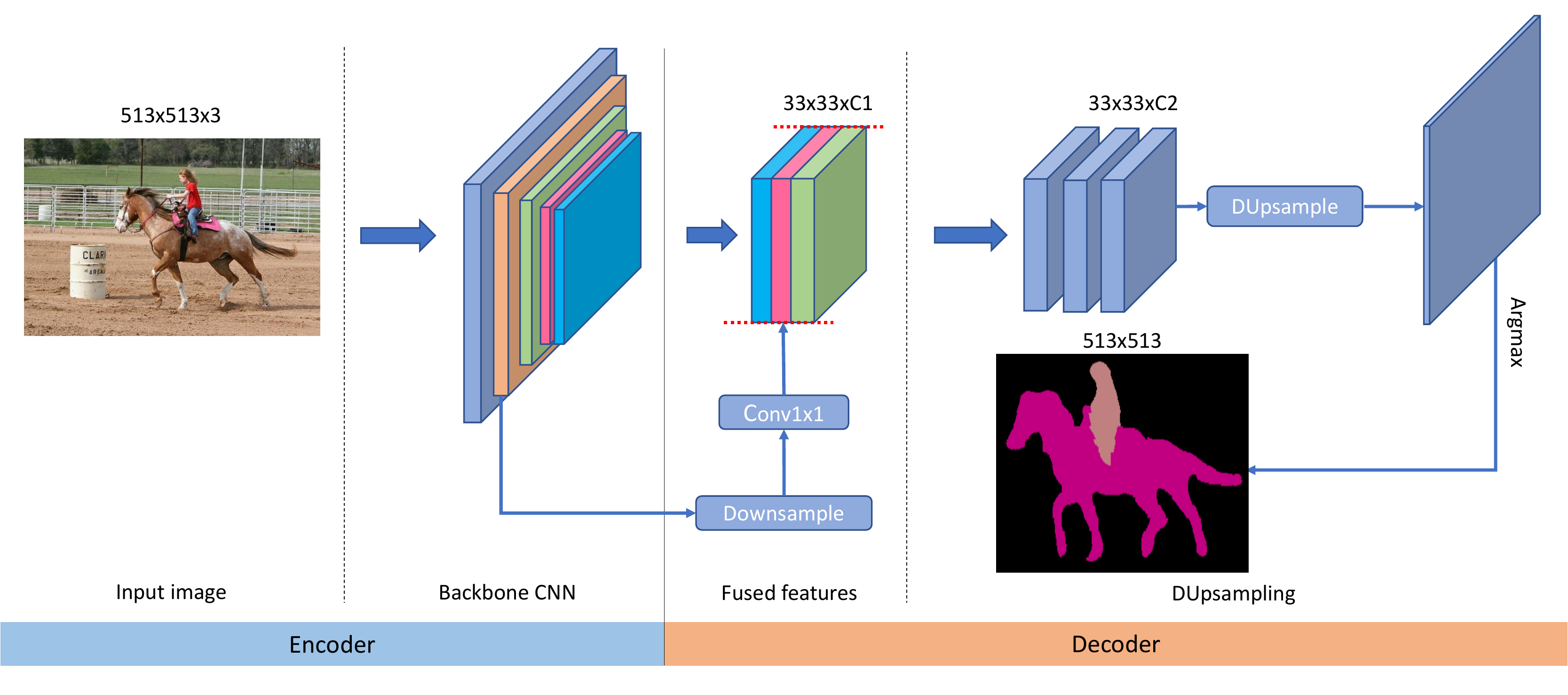}
  \caption{The framework with our proposed decoder. The major differences from the previous framework shown in Fig.~\ref{fig:vanilla_decoder} are 1) all fused features are downsampled to the lowest features resolution before merging. 2) The incapble bilinear is replaced with our proposed DUpsampling to recover the full-resolution prediction. }
  \vspace{-0.5cm}
  \label{fig:main_figure}
\end{figure*}

Fully convolutional networks (FCNs) \cite{long2015fully} have achieved tremendous success in dense pixel prediction applications such as  semantic segmentation, for which the algorithm is asked to predict a variable for each pixel of an input image and is a fundamental problem in computer vision. The great achievement of FCNs results from powerful features extracted by  CNNs. Importantly, the sharing convolutional computation mechanism makes training and inference computationally very efficient.

In the original FCNs, several stages of strided convolutions and$/$or spatial
pooling reduce the final image
prediction typically by a factor of $32$, thus losing fine image structure information and leading to inaccurate predictions, especially at the object boundaries.
DeepLab \cite{chen2018deeplab} applies atrous (a.k.a dilation) convolutions, achieving large receptive fields while maintaining a higher-resolution feature map.
Alternatively the encoder-decoder architecture is often used to  address this  problem. The encoder-decoder architecture views the backbone CNN as an encoder, responsible for encoding a raw input image into lower-resolution feature maps (e.g., $\frac{1}{r}$ of the input image size with  $r=8,16,$ or $32$). Afterwards, a decoder is used to recover the pixel-wise prediction from the lower-resolution feature maps. In previous works \cite{chen2018encoder, lin2017refinenet}, a decoder consists of %
a few convolutional layers
and a bilinear upsampling. The light-weight convolutional decoder yields high-resolution feature maps and  bilinear upsampling  is finally applied to the resulting feature maps to obtain the desired pixel-wise prediction. The decoder commonly fuses low-level features to capture the fine-grained information lost by convolution and pooling operations in CNNs. A standard DeepLabv3+ encoder-decoder architecture is illustrated in Fig.~\ref{fig:vanilla_decoder}.

A drawback of the oversimple bilinear upsampling  is  its
limited capability
in recovering the pixel-wise prediction accurately.
Bilinear upsampling
does not take into account the correlation among the prediction of each pixel since it is data independent.
As a consequence,  the convolutional decoder is required to produce
relatively higher-resolution feature maps in order to obtain good final prediction (e.g., $ \frac{1}{4} $ or $ \frac{1}{8}$ of the input size).
This requirement causes two
issues for semantic segmentation.

1) The overall strides of the encode  must be reduced
very aggressively by using
 multiple atrous convolutions \cite{chen2018deeplab, yu2015multi}.
 The price is much heavier
  computation complexity and memory footprint,
  hampering the training on large data and deployment for real-time applications.

  For example, in order to achieve  state-of-the-art performance, the recent DeepLabv3+ \cite{chen2018encoder} reduces the overall strides of its encoder by four times (from 32 to 8). Thus  inference of DeepLabv3+ is very slow.
  2) The decoder is often needed  to fuse features at  very low levels. For example, DeepLabv3+ fuses features of ${\sf downsample \; ratio} = 4$\footnote{$\sf downsample \; ratio$ denotes the ratio of the resolution of the feature maps to that of the input image.} in
  $\sf block1$ as shown in Fig.~\ref{fig:vanilla_decoder}. It is because that the fineness of the final prediction is actually dominated by the resolution of the fused low-level features due to the inability of bilinear. As a result, in order to produce high-resolution prediction, the decoder has to fuse the high-resolution features at a low level. This constraint narrows down the design space of the feature aggregation and therefore is likely to cause a {\it suboptimal} combination of features to be aggregated  in the decoder. In experiments, {\em we show that a better feature aggregation strategy can be found if the feature aggregation can be designed without the constraint imposed by the resolution of feature maps.
}

In order to tackle the aforementioned issues caused by bilinear, here we propose a new data-dependent upsampling method, termed DUpsamling,
to recover the pixel-wise prediction from the final outputs of the CNNs, replacing bilinear upsampling used extensively in previous works. Our proposed DUpsampling takes advantages of the redundancy in the segmentation label space and proves to be capable of accurately recovering the pixel-wise prediction from relatively coarse CNNs outputs, alleviating the need for precise responses from the convolutional decoder.

{\em As a result, the encoder is not required to overly reduce its overall strides, dramatically reducing the computation time and memory footprint of the whole segmentation framework}. Meanwhile, due to the effectiveness of DUpsampling, it allows the decoder to downsample the fused features to the lowest resolution of feature maps before merging them. This downsampling not only reduces the amount of computation of the decoder, but much more importantly it decouples the resolution of fused features and that of the final prediction. This decoupling allows the decoder to make use of arbitrary  feature aggregation and thus a better feature aggregation can be leveraged so as to boost the  segmentation performance as much as possible.

Finally, DUpsampling can be seamlessly incorporated into the network with a standard 1$\times$1 convolution and thus needs no ad-hoc coding. Our overall framework is shown in Fig.~\ref{fig:main_figure}.

We summarize our main contributions as follows.
\begin{itemize}
\itemsep -.05112cm
    \item We propose a simple yet effective Data-dependent Upsampling (DUpsampling) method to recover the pixel-wise segmentation prediction from the coarse outputs of the convolutional decoder,
    replacing the incapable bilinear used extensively in previous methods.
    \item Taking advantages of our proposed DUpsampling, we can avoid overly reducing the overall strides of the encoder, significantly reducing the computation time and memory footprint of
    the semantic segmentation method  by a factor of 3 or so.
    \item DUpsampling also allows the decoder to downsample the fused features to the lowest resolution of feature maps before merging them. The downsampling not only reduces the amount of computation of the decoder dramatically but also enlarges the design space of feature aggregation, allowing a better feature aggregation to be exploited in the decoder.
    \item Together with the above contributions, we propose a new decoder scheme, which compares favourably with state-of-the-art decoders while using $\sim$20\% amount of computation. With the proposed decoder, the framework illustrated in Fig.~\ref{fig:main_figure} achieves new state-of-the-art performance: mIOU of 88.1\%\footnote{The results on PASCAL VOC {\it test} set can be found at \url{http://host.robots.ox.ac.uk:8080/anonymous/UYT221.html}
    } on PASCAL VOC \cite{everingham2010pascal} with only  30\% computation of the previous %
    best
    framework of DeepLabv3+ \cite{chen2018encoder}.
    We also set a new mIOU record of 52.5\% on the PASCAL Context dataset
    \cite{mottaghi2014role}.
\end{itemize}

\section{Related Work}

Efforts have been devoted  to improve  pixel-wise predictions with FCNs. They can be roughly divided into two groups:
atrous convolution \cite{chen2018deeplab, yu2015multi} and encoder-decoder architectures \cite{lin2017refinenet, chen2018encoder, long2015fully, badrinarayanan2017segnet, noh2015learning}.

\textbf{Atrous convolution.} A straightforward approach
is to reduce the overall strides of backbone CNNs by dropping some strided  convolutions or pooling layers.  However, simply reducing these strides would diminish the receptive field of convolution networks substantially, which proves to be crucial to semantic segmentation \cite{chen2018deeplab, peng2017large, liu2015parsenet}.
Atrous convolutions   \cite{chen2017rethinking, chen2018deeplab, chen2018encoder, yu2015multi} can be used to keep the receptive field unchanged, meanwhile not downsampling the feature map resolution too much.
The major drawback of atrous convolutions is much heavier computation complexity and larger memory requirement as the size of those atrous convolutional kernels, as well as the resulted feature maps, become much larger
 \cite{he2016deep, chollet2017xception}.

\textbf{Encoder-decoder architectures.} Encoder-decoder architectures are proposed to overcome the drawback of atrous convolutions and are widely used  for semantic segmentation. DeconvNet \cite{noh2015learning}  uses stacked deconvolutional layers to recover the full-resolution prediction gradually. The method has the potential to produce high-resolution prediction but is difficult to train
due to many parameters introduced by the decoder. SegNet \cite{badrinarayanan2017segnet} shares a similar idea with DeconvNet but uses indices in pooling layers to guide the recovery process,  resulting in better performance. RefineNet \cite{lin2017refinenet}
further fuse low-level features to improve the performance. Recently, DeepLabv3+ \cite{chen2018encoder} takes advantages of both encoder-decoder architectures and atrous convolution, achieving best reported performance on a few datasets to date. Although efforts have been spent on designing a better decoder, so far almost none of them can bypass the restriction on the resolutions of the fused features and exploit better feature aggregation.

\section{Our Approach}

In this section, we firstly reformulate semantic segmentation with our proposed DUpsampling and then present the adaptive-temperature softmax function
which makes the training with DUsampling much easier.
Finally, we show how the framework can be largely improved with the fusion of downsampled low-level features.

\subsection{Beyond Bilinear: Data-dependent Upsampling} \label{section:dupsampling}
\begin{figure*}[t]
  \centering
  \includegraphics[width=.8\linewidth]{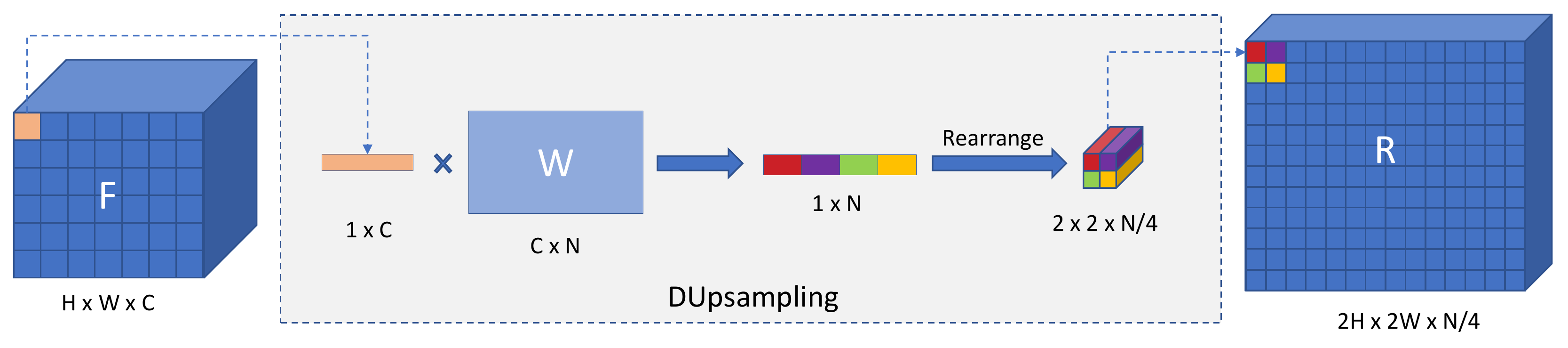}
  \caption{The proposed DUpsampling. In the figure, DUpsampling is used to upsample the CNNs outputs $\mF$  by twice. $\mR$ denotes the resulting maps. $\mW$, computed with the method described in Sec.~\ref{section:dupsampling}, is the inverse projection matrix of DUpsampling. In practice, the upsampling ratio is typically 16 or 32.}
  \vspace{-0.2cm}
  \label{fig:dupsampling}
  \vspace{-0.2cm}
\end{figure*}
In this section, we firstly consider the simplest decoder, which is only composed of upsampling. Let $\mF \in \Real^{\tilde{H} \times \tilde{W} \times \tilde{C}}$ denote the final outputs of the encoder CNNs and $ \mY \in \{0, 1, 2, ..., C\}^{H \times W}$ be the ground truth label map, where $C$ and $\tilde{C}$ denotes the number of classes of segmentation and the number of channels of the final outputs, respectively.
$\mY $ is commonly encoded with one-hot encoding, i.e., $\mY \in \{0, 1\}^{H \times W \times C}$. Note that $\mF$ is typically of a factor of  16 or 32  in spatial size of the  ground-truth $\mY$.  In other words, $ \frac{\tilde{H}}{H} = \frac{\tilde{W}}{W} = \frac{1}{16}$ or $\frac{1}{32}$. Since semantic segmentation requires per-pixel prediction, $\mF$ needs to be upsampled to the spatial size of $\mY$ before computing the training loss.

Typically in semantic segmentation  \cite{chen2017rethinking, chen2018encoder, long2015fully, zhang2018context, he2019knowledge}, the training loss function is formulated as:
\begin{equation} \label{bilinear_upsampling_loss}
{\sf L}(\mF, \mY) = {\Loss}(\softmax(\bilinear(\mF)), \mY)).
\end{equation}
Here $\Loss$ is often the cross-entropy loss,  and $\bilinear$
is used to upsample $\mF$ to the spatial size of $\mY$.
We argue that bilinear unsampling may not be the optimal choice here.
As we  show in the experiments (Sec.~\ref{section:dupsampling_vs_bilinear}), bilinear is oversimple and has an inferior  upper bound in terms of  reconstructing (best possible reconstruction quality). In order to
compensate
the loss caused by bilinear, the employed deep network is consequently required to
output  higher-resolution feature maps, which are input to the bilinear operator.
As mentioned above, the solution is to apply atrous convolutions, with the price of high computation complexity.
For example, {\em  reducing the overall strides from $16$ to $8$ incurs more than 3 times computation.}

An important observation is that the semantic segmentation label $\mY$
of an image is not i.i.d.\ and there contains structure information
so that $\mY$ can be compressed considerably, with almost no loss.
Therefore, unlike  previous methods, which upsample $\mF$ to the spatial size of $\mY$, we instead attempt to compress $\mY$ into $ \tilde{\mY} \in \Real^{\tilde{H} \times \tilde{W} \times \tilde{C}}$ and then compute the training loss between $\mF$ and $\tilde{\mY}$. Note that $\mF$ and $\tilde{\mY}$ are of  the same  size.

In order to compress $\mY$ into $\tilde{\mY}$,  we
seek a transform under some metric to minimize the reconstruction error between $ \mY  $ and  $\tilde{\mY}$.
Specifically, let $r$ indicate the ratio of $H$ to $\tilde{H}$, which is usually $16$ or $32$.
Next, $\mY$ is divided into an ${\frac{H}{r}} \times \frac{W}{r}$ grid of sub-windows of size $r \times r$ (if $H$ or $W$ is not dividable by $r$, a padding is applied).
For each sub-window $ \mS   \in \{0, 1\}^{r \times r \times C}$, we reshape $ \mS $ into a vector ${\boldsymbol{v}} \in \{0, 1\}^N$, with  $N = r \times r \times C$.
Finally, we compress the vector  $\boldsymbol{v}$ to a lower-dimensional  vector  $\boldsymbol{x} \in \Real^{ \tilde C }$ and then vertically and horizontally stack all $\boldsymbol{x}$'s to form $\tilde{\mY}$.

Although a variety of ways can be used to achieve the compression, we find that simply using linear projecting, i.e.,   multiplying $ \vv $ by a matrix $\mP  \in \Real^{\tilde{C} \times N}$ works well in this case. Formally, we have,
\begin{equation}
\boldsymbol{x} = \mP  \boldsymbol{v}; \;\;\;  \boldsymbol{\tilde{v}} = \mW\boldsymbol{x},
\end{equation}
where $\mP  \in \Real ^{\tilde{C} \times N}$ is used to compress $\boldsymbol{v}$ into $\boldsymbol{x}$. $ \mW \in \R^{N \times \tilde{C}}$ is the inverse projection matrix (a.k.a.\ reconstruction matrix) and used to reconstruct  $\boldsymbol{x}$ back to  $\boldsymbol{v}$. $\boldsymbol{\tilde{v}}$ is  the reconstructed $\boldsymbol{v}$. We have omitted the offset term here. In practice  prior to the compression, $\boldsymbol{v}$ is centered by subtracting its mean over the training set.

The matrices $\mP $ and $\mW$ can be found by minimizing the reconstruction error between $\boldsymbol{v}$ and $\boldsymbol{\tilde{v}}$ over the  training set. Formally,
\begin{equation}
\begin{aligned}
\mP ^*, \mW^* = \underset{\mP , \mW}{\arg\min} \sum_{\boldsymbol{v}} || \boldsymbol{v} - \boldsymbol{\tilde{v}} ||^2 \\
= \underset{ \mP, \mW}{\arg\min} \sum_{\boldsymbol{v}} || \boldsymbol{v} - \mW \mP \boldsymbol{v} ||^2.
\end{aligned}
\end{equation}
This objective can be iteratively optimized with standard stochastic gradient descent (SGD). With an  orthogonality constraint, we can simply use principal component analysis (PCA) \cite{wold1987principal} to achieve a closed-form solution for the objective.

Using $\tilde{\mY}$ as the target,
we may pre-train the network with a regression loss by observing that
the compressed labels $ \tilde \mY $ is real-valued
\begin{equation} \label{dupsampling_loss}
{\sf L}( \mF, \mY) = ||\mF - \tilde{\mY}||^2.
\end{equation}
Thus any regression loss, $ \ell_2 $ being a typical example as in Eq.~\eqref{dupsampling_loss},    can be employed here.
Alternatively,
a more direct approach is to
 compute the loss in the space of $\mY$. Therefore, instead of compressing $\mY$ into $\tilde{\mY}$, we up-sample $\mF$  with the learned reconstruction  matrix $ \mW   $   and then compute the pixel classification loss between the decompressed $\mF$ and $\mY$:
\begin{equation} \label{dupsampling_dual_loss}
{\sf L}(\mF, \mY) = \Loss(\softmax( {\sf DUpsample} ( \mF )  ), \mY).
\end{equation}
With linear reconstruction, $ {\sf DUpsample} ( \mF )$  applies linear upsampling of $ \mW \vf   $ to each feature $ \vf \in \Real^{ \tilde{C}} $ in the tensor $ \mF $.
Comparing with Eq.~\eqref{bilinear_upsampling_loss},
we have replaced the bilinear upsampling with a data-dependent upsampling, learned from the ground-truth labels.
This upsampling procedure is essentially the same as applying a $ 1$$\times$$ 1$ convolution
along the spatial dimensions, with convolutional kernels stored in $ \mW $.
The decompression is illustrated in Fig.~\ref{fig:dupsampling}.

Note that, besides the linear upsampling presented above, we have also
conducted experiments using a nonlinear auto-encoder for upsampling.
Training of the auto-encoder is also to minimize the reconstruction loss,
and is  more general than the linear case. Empirically,
we observe that the final semantic prediction accuracy is almost the same
as using the much simpler linear reconstruction. Therefore we
focus on using the linear reconstruction in the sequel.

\textbf{Discussion with Depth-to-Space and Sub-pixel.} The simplest linear form of DUpsample can be viewed as an improved version of Depth-to-Space in \cite{wojna2017devil} or Sub-pixel in \cite{shi2016real} with pre-computed upsampling filters.
Depth-to-Space and Sub-pixel are typically used to upsample the inputs by a modest upsample ratio (e.g., $\leq 4$), in order to avoid incurring too many trainable parameters resulting in difficulties in optimization.
In contrast, as the upsampling filters in our method are pre-computed, the upsample ratio of DUpsamling can be very large (e.g., $16$ or $32$) if needed.

\subsection{Incorporating DUpsampling with Adaptive-temperature Softmax}

So far, we have shown that DUpsampling can be used to replace the incapable bilinear upsampling in semantic segmentation. The next step is to incorporate the DUpsampling into the encoder-decoder framework, resulting in an end-to-end trainable system. While DUpsampling can be realized with a 1$\times$1 convolution operation, incorporating directly it into the framework encounters difficulties in optimization.

Probably due to the $\mW$ is computed with one-hot encoded $\mY$, we find that the combination of vanilla softmax and DUpsampling has difficulty in producing sharp enough activation. As a result, the cross-entroy loss is stuck during the training process (as shown in experiment \ref{section:temperature_softmax}), which makes the training process slow to converge.
In order to tackle the issue, we instead employ the softmax function with temperature \cite{hinton2015distilling}, which adds a temperature $T$ into vanilla softmax function to sharpen/soften the activation of softmax.
\begin{equation}
\softmax(z_i) = \frac{\exp({z_i/T})}{\sum_{j} \exp(z_j/T)}.
\end{equation}

We find that $T$ can be learned automatically using the standard back-propagation algorithm, eliminating the need for tuning. We show in experiments that this adaptive-temperature softmax makes training converge much faster without introducing extra hyper-parameters.
\subsection{Flexible Aggregation of Convolutional Features}
The extremely deep CNNs \cite{he2016deep, chollet2017xception, huang2017densely} lead to the success in computer vision. However, the depth also causes the loss of fine-grained information essential to semantic segmentation. It has been shown by a number of works \cite{lin2017refinenet, chen2018encoder} that combining the low-level convolutional features can improve the segmentation performance significantly.

Let $\mF$ be the eventual CNNs feature maps used to produce the final pixel-wise prediction by bilinear or aforementioned DUpsampling. $\mF_i$ and $\mF_{\rm last}$ represent the feature maps at level $i$ of the backbone and last convolutional feature maps of the backbone, respectively.  For simplicity we focus on fusing one level of low-level features, but it is straightforward to extend it to multi-level fusion, which perhaps boosts the performance further. The feature aggregation in previous decoders shown in Fig.~\ref{fig:vanilla_decoder} can be formulated as,
\begin{equation}
\label{traditional_features_fusion}
    \mF = f(\concat(\mF_i, \upsample(\mF_{\rm last}))),
\end{equation}
where $ f $ denotes a CNN and $\upsample$ is usually bilinear.
$ \concat $ is a concatenation operator  along the channel. As described above, this arrangement comes with two problems. 1)  $f$ is applied after upsampling. Since $f$ is a CNN, whose amount of computation depends on the spatial size of inputs, this arrangement would render the decoder inefficient computationally. Moreover, the computational overhead prevents the decoder from exploiting features at a very low level. 2) The resolution of fused low-level features $\mF_i$ is equivalent to that of $\mF$,
which is typically around $ \frac{1}{4} $  resolution of the final prediction due to the incapable bilinear used to produce the final pixel-wise prediction. In order to obtain high-resolution prediction, the decoder can only choose the feature aggregation with high-resolution low-level features.

In contrast, in our proposed framework, the  responsibility to restore the full-resolution prediction has been largely shifted to DUpsampling. Therefore, we can safely downsample  any level of used low-level features to the resolution of last feature maps $F_{\rm last}$ (the lowest resolution of feature maps) and then fuse these features to produce final prediction, as shown in Fig.~\ref{fig:main_figure}. Formally, Eq.~\eqref{traditional_features_fusion} is changed to,
\begin{equation} \label{proposed_features_fusion}
\mF = f(
\concat(\downsample(\mF_i), \mF_{\rm last})),
\end{equation}
where $\downsample$ is bilinear in our case. This rearrangement not only keeps the features always to be computed efficiently at the lowest resolution, but also decouples the resolution of low-level features $\mF_i$ and that of the final segmentation prediction, allowing any level of features to be fused. In experiments, we show the flexible feature fusion enables us to exploit a better feature fusion to boost the segmentation performance as much as possible.

Only when cooperating with the aforementioned DUpsampling, the scheme of downsampling low-level features can work. Otherwise, the performance is bounded by the upper bound of the incapable upsampling method of the decoder.
This is the reason why previous methods are required to upsample the low-resolution high-level feature maps back to the spatial size of fused low-level feature maps.

\section{Experiments}

The proposed models are evaluated on the PASCAL VOC 2012 semantic segmentation benchmark \cite{everingham2010pascal} and PASCAL Context benchmark \cite{mottaghi2014role}. For both benchmarks, we measure the performance in terms of pixel intersection-over-union averaged across the present classes (i.e., mIOU).

\textbf{PASCAL VOC} is the  dataset widely used for semantic segmentation. It consists of 21 classes including background. The splits of PASCAL VOC are $1, 464$, $1, 449$ and $1, 456$ for training, validation and test, respectively. The ablation study of our work is conducted over its {\it val} set. Also, we report our performance over {\it test} set to compare with other state-of-the-art methods.

\textbf{PASCAL Context} is much larger than PASCAL VOC,
including $4, 998$ images for training and $5, 105$ images for validation. Following previous works \cite{lin2017refinenet, mottaghi2014role}, we choose the most frequent 59 classes plus one background class (i.e., 60 classes in total) in our experiments. There is not a test server available and therefore we follow previous works \cite{lin2017refinenet, zhang2018context, chen2018deeplab, long2015fully, zheng2015conditional} to report our result on {\it val} set.

\textbf{Cityscapes} is a large-scale benchmark for semantic urban scene parsing. It contains $2, 975$ images for training, 500 images for validation and $1, 525$ images for testing. Additionally, it also provides about $20,000$ weakly annotated images.

\textbf{Implementation details.} For all ablation experiments on PASCAL VOC, we opt for ResNet-50 \cite{he2016deep} and Xception-65 \cite{chollet2017xception} as our backbone networks, both of which are modified as in \cite{chen2018encoder}. Following \cite{liu2015parsenet, chen2017rethinking, chen2018encoder}, we use ``poly" as our learning rate policy for all experiments. The initial learning rate is set as 0.007 and total iteration is $30k$ for ablation experiments on PASCAL VOC. For all ResNet-based experiments, weight decay is set to 0.0001. The batch size is set to 48, but the batch normalization \cite{ioffe2015batch} statistics are computed with a batch of 12 images. For all Xception-based experiments, weight decay is 0.00004.
We use a batch size of 32 but compute the batch normalization statistics within a batch of 16 images. We follow the  practice \cite{chen2018encoder, chen2017rethinking, zhang2019deep} to use the weights pre-trained on ImageNet \cite{deng2009imagenet} to initialize backbone networks. All weights of newly added layers are initialized with Gaussian distribution of variance 0.01 and mean 0. $T$ in adaptive-temperature softmax is initialized to 1. $\tilde{C}$ is set as 64 for ResNet-50 based experiments and 128 for Xception-65 based experiments. Finally, following previous works \cite{chen2017rethinking, chen2018deeplab, chen2018encoder}, we augment the training data by randomly scaling the images from 0.5 to 2.0 and left-right flipping them.

\subsection{Ablation Study}
Our work focuses on the decoder part of the segmentation architecture. Therefore, for all ablation experiments, we use the same encoder, as shown in Fig. \ref{fig:vanilla_decoder}. The encoder yields the final feature maps with the $\frac{1}{ 16}$ or $\frac{1}{32}$ size of the original image. The decoder aims to decode the low-resolution feature maps into the prediction with the same resolution as the original image. In this section, we will investigate different decoder schemes, and demonstrate our proposed decoder's advantages. We make use of official {\it train} set instead of SBD \cite{hariharan2011semantic} since it provides more consistent annotations.

\subsubsection{DUpsampling vs.\  Bilinear} \label{section:dupsampling_vs_bilinear}
First of all, we design experiments to show that the upper bound of bilinear is much lower than that of DUpsampling, which results in limited performance of bilinear. Specifically, we design a light-weight CNN including five convolutional layers with kernel size being 3 and stride of 2, which is fed  with {\em ground truth labels} instead of raw images. Next, DUpsampling or bilinear is added on top of that to recover the pixel-wise prediction.
This is similar to the {\em decoder} part in the encoder-decoder architecture.

By training the two networks, with DUpsampling or bilinear as decoder respectively, the ability to restore the pixel-wise prediction can be quantitatively measured via their performance over the {\it val} set, which can be viewed as the upper bound of both methods. We use the training protocol described in implementation details to train the two networks, except that the total iterations and initial learning rate are set as $100k$ and 0.07, respectively.  ``output stride'' indicates the ratio of input image spatial resolution to the final CNN feature maps resolution. As shown in Table~\ref{table:pca_vs_bilinear}, the upper bound performance of DUpsampling is well above that of bilinear both when output stride being 32 and  16.

Given the superior upper bound performance of DUpsampling, we further carry out experiments with raw input images. In the experiments, we employ ResNet-50 as the backbone network. Unsurprisingly, by merely replacing the bilinear with DUpsampling, the mIOU on PASCAL VOC {\it val} set is improved by 1.3 points and 1 point, when the output stride is 32 and 16 respectively, as shown in Table \ref{table:pca_vs_bilinear}. The improvement is significant because mIOU is strict. Interestingly, the DUpsampling of output stride being 32 achieves similar performance to the bilinear case of output stride being 16.
This shows that the proposed DUpsampling may eliminate the need for expensive computationally high-resolution feature maps from the  CNNs.

\begin{table}
\begin{center}
\small
\begin{tabular}{ l|c|c|c  }
\hline
Method & output stride & mIOU (\%) & mIOU* (\%)\\
\hline\hline
bilinear & 32 & 70.77 & 94.80\\
DUpsampling & 32 & 72.09 & 99.90 \\
bilinear & 16 & 72.15 & 98.40 \\
DUpsampling & 16 & \textbf{73.15} & \textbf{99.95} \\
\hline
\end{tabular}
\end{center}
\vspace{-0.3cm}
\caption{mIOU over the PASCAL VOC $val$ set of DUpsampling vs.\ bilinear upsampling.  ``output stride'' indicates the ratio of input image spatial resolution to final output resolution. mIOU* denotes the upper bound.}
\label{table:pca_vs_bilinear}
\vspace{-0.5cm}
\end{table}

\subsubsection{Flexible aggregation of convolutional features}

\begin{table}
\begin{center}
\small
\begin{tabular}{l|c|c}
\hline
Used low-level features & mIOU (\%) & FLOPS\\
\hline\hline
N/A & 73.15 & 0.80B \\
conv1\_3 & 72.70 & 1.13B \\
b1u2c3 & 74.03 & 1.15B \\
b3u6c3 & 73.43 & 1.23B \\
b1u2c3 + b3u6c3 & 73.82 & 1.58B \\
conv1\_3 + b3u6c3 & \textbf{74.20} & 1.56B \\
\hline
\end{tabular}
\end{center}
\vspace{-0.5cm}
\caption{mIOU over PASCAL VOC $val$ set when using different fusion of features. b$x$u$y$c$z$ denotes low-level features named block\_$x$/unit\_$y$/conv\_$z$ in ResNet. "FLOPS" denotes the amount of computation of the decoder including feature aggregation, convolutional decoder and the final upsampling.}
\label{table:combination_of_features}
\vspace{-0.7cm}
\end{table}

\begin{figure}[!b]
  \centering
  \includegraphics[width=\linewidth]{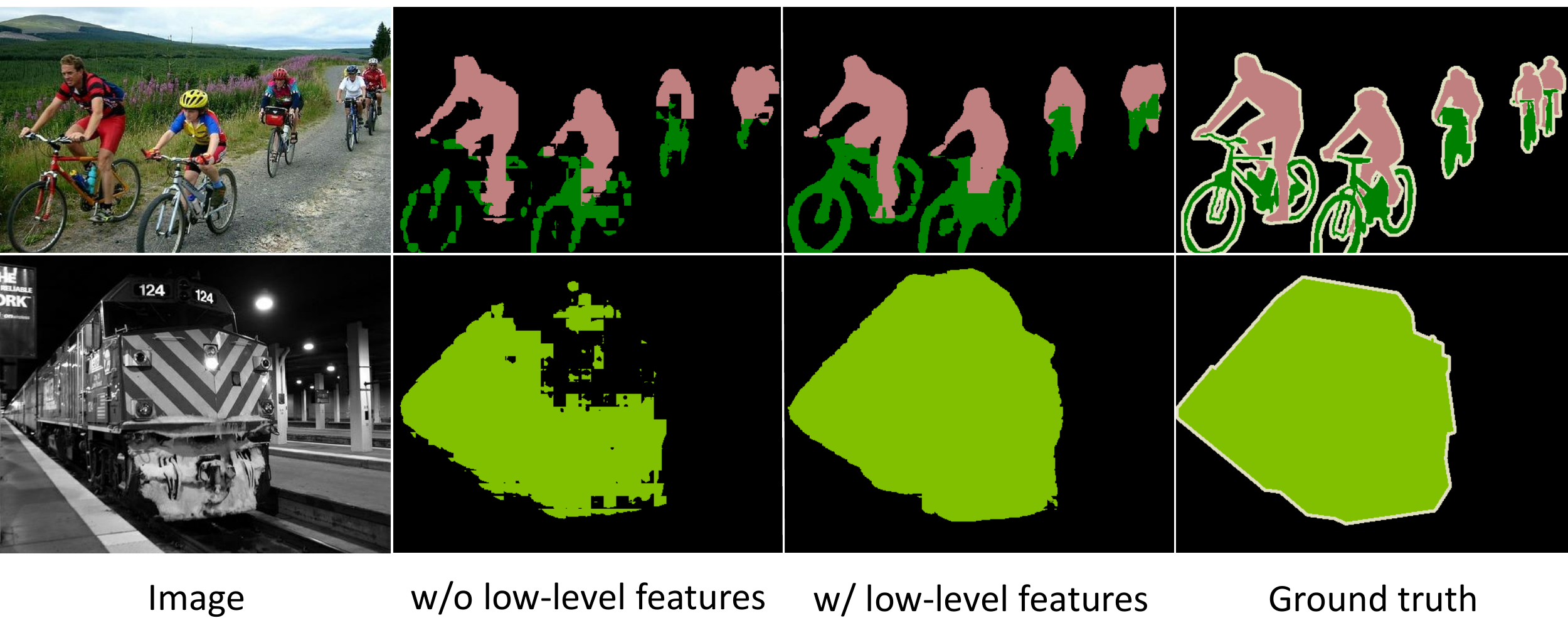}
  \caption{The prediction results with low-level features and without low-level features. ResNet-50 is used as the backbone.}
  \label{fig:prediction_results}
\end{figure}

Due  to the flexibility of our proposed decoder, we can employ any combination of features to improve segmentation performance, regardless of the resolution of fused features.

For ResNet-50, we experiment with many different combinations of features, as shown in Table~\ref{table:combination_of_features}. The best one is the combination of conv1\_3 + b3u6u3, achieving mIOU 74.20\% over $val$ set. Additionally, as shown in Table \ref{table:combination_of_features}, the amount of computation changes little when features at different levels are fused, which allows us to choose the best feature fusion without considering the price of computation incurred by the resolution of fused features.

In order to understand how the fusion works, we visualize the segmentation results with and without low-level features in Fig. \ref{fig:prediction_results}. Intuitively, the one fusing low-level features yields more consistent segmentation, which suggests the downsampled low-level features are still able to refine the segmentation prediction substantially.

\subsubsection{Comparison with the vanilla bilinear decoder}
\begin{table}
\begin{center}
\small
\begin{tabular}{  l|l|c|c  }
\hline
Decoder & Low-level features / ratio & mIOU (\%) & FLOPS \\
\hline\hline
\multicolumn{4}{ c }{ResNet-50} \\
\hline
Vanilla & b1u2c3 / 4 & 73.26 & 5.53B \\
Proposed & b1u2c3 / 4 & 74.03 & 1.15B \\
Vanilla & conv1\_3 / 2 + b3u6c3 / 16 & - & 22.34B \\
Proposed & conv1\_3 / 2 + b3u6c3 / 16 & \textbf{74.20} & 1.56B \\
\hline\hline
\multicolumn{4}{  c }{Xception-65} \\
\hline
Vanilla & efb2u1c2 / 4 & 78.70 & 5.53B \\
Proposed & efb2u1c2 / 4 & 79.09 & 1.93B \\
Vanilla & mfb1u16c3 / 16 & 78.74 & 0.41B \\
Proposed & mfb1u16c3 / 16 & \textbf{79.67} & 1.98B \\
\hline
\end{tabular}
\end{center}
\vspace{-0.4cm}
\caption{mIOU over the PASCAL VOC $val$ set when using different fusion strategies of features. b$x$u$y$c$z$ denotes low-level features named block\_$x$/unit\_$y$/conv\_$z$ in ResNet or Xception. ``ef" and ``mf" respectively indicate ``entry\_flow" and ``middle\_flow" in Xception. ``-" means out-of-memory. ``ratio" denotes the ratio of the resolution of feature maps to the resolution of the input image (i.e., ${\sf downsample\; ratio}$). ``FLOPS" denotes the amount of computation of the decoders.}
\label{table:comparison_decoders}
\end{table}

\begin{table}
\begin{center}
\small
\begin{tabular}{  l|l|c|c  }
\hline
Decoder & low-level features / ratio & mIOU (\%) & FLOPS \\
\hline\hline
Vanilla & efb2u1c2 / 4 & \textbf{79.36} & 43.65B \\
Proposed & mfb1u16c3 / 16 & 79.06 & 25.14B \\
\hline
\end{tabular}
\end{center}
\caption{mIOU on the Cityscapes $val$ set. Our proposed decoder with much less computation complexity achieves a similar performance as the vanilla decoder.}
\label{table:comparison_decoders_cityscapes}
\end{table}

We further compare our proposed decoder scheme with the vanilla bilinear decoder shown in Fig. \ref{fig:vanilla_decoder}, which fuses low-level features b1u2c3 (${\sf downsample\; ratio} = 4$). As shown in Table \ref{table:comparison_decoders}, it achieves mIOU 73.26\% on {\it val} set with ResNet-50 as the backbone. By replacing vanilla decoder with our proposed decoder in Fig. \ref{fig:main_figure}, the performance is improved to 74.03\%. Because of the same low-level features used, the improvement should be due to the capable DUpsampling instead of bilinear used to restore the full-resolution prediction. Furthermore, we explore a better feature fusion conv1\_3 + b3u6c3 for proposed deocder and improve the overall performance slightly to 74.20\%. When the vanilla decoder uses the fusion of features, it incurs much heavier computation computation complexity and runs out of our GPUs memory due to the high resolution of conv1\_3, which prevents the vanilla decoder from exploiting the low-level features.

We also experiment our proposed decoder with Xception-65 as the backbone. Similarly, with the same low-level features efb2u1c3 (${\sf downsample\; ratio} = 4$), our proposed decoder improves the performance from 78.70\% to 79.09\%, as shown in Table \ref{table:comparison_decoders}. When using a better low-level features mfb1u16c3 (${\sf downsample\; ratio} = 16$), the vanilla decoder just improves the performance negligibly by 0.04\% because its performance is constrained by the incapable bilinear upsampling used to restore the full-resolution prediction. In contrast, our proposed decoder can still benefit a lot from the better feature fusion due to the use of much powerful DUpsampling. As shown in Table \ref{table:comparison_decoders}, with the better feature fusion, the performance of our proposed decoder is improved to 79.67\%. Moreover, since we downsample low-level features before fusing, our proposed decoder requires much fewer FLOPS than the vanilla decoder of the best performance, as shown in Table \ref{table:comparison_decoders}.

Finally, we compare our proposed decoder with the vanilla bilinear decoder on the Cityscapes $val$ set. Following \cite{chen2018encoder}, Xception-71 is used as our backbone and the number of iterations is increased to $90k$ with a initial learning rate being 0.01. As shown in Table \ref{table:comparison_decoders_cityscapes}, under the same training and testing settings, our proposed decoder achieves a comparable performance with the vanilla one while using much less  computation.
\subsubsection{Impact of adaptive-temperature softmax} \label{section:temperature_softmax}
\begin{figure}[ht!]
  \centering
  \includegraphics[width=.9800976\linewidth]{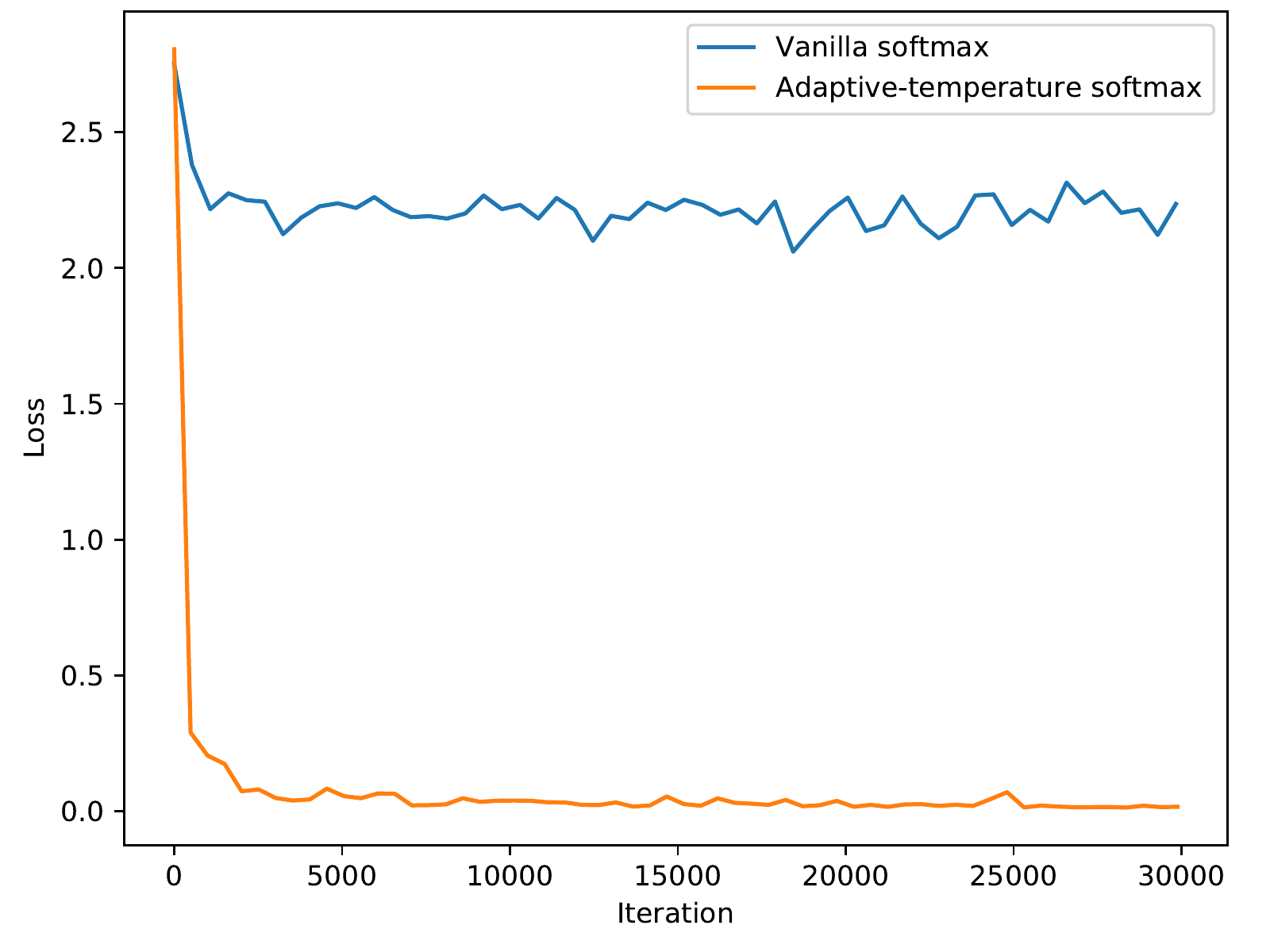}
  \caption{Training losses for vanilla softmax and adaptive-temperature softmax.}
  \label{fig:softmax_losses}
\end{figure}

As mentioned before, the adaptive-temperature softmax eases the training of the proposed DUpsampling method. When training the framework with vanilla softmax with $T$ being 1, it achieves 69.81\% over {\it val} set, which is significantly lower than 73.15\% of the counterpart with adaptive-temperature softmax. We further plot training losses for vanilla softmax and adaptive-temperature softmax in Fig. \ref{fig:softmax_losses}, which shows the advantage of this adaptive-temperature softmax.

\subsection{Comparison with state-of-the-art Methods}
\begin{table}
\begin{center}
\small
\begin{tabular}{ r |c}
\hline
Method & mIOU (\%) \\
\hline\hline
PSPNet \cite{zhao2017pyramid} & 85.4 \\
DeepLabv3 \cite{chen2017rethinking} & 85.7 \\
EncNet \cite{zhang2018context} & 85.9 \\
DFN \cite{yu2018learning} & 86.2 \\
IDW-CNN \cite{wang2017learning} & 86.3 \\
CASIA\_IVA\_SDN \cite{fu2017stacked} & 86.6 \\
DIS \cite{luo2017deep} & 86.8 \\
\hline
DeepLabv3+ \cite{chen2018encoder} (Xception-65) & 87.8 \\
Our proposed (Xception-65) & \textbf{88.1} \\
\hline
\end{tabular}
\end{center}
\caption{State-of-the-art methods on PASCAL VOC {\it test} set.}
\label{table:state_of_the_art_methods}
\end{table}
\begin{table}
\begin{center}
\small
\begin{tabular}{ r |c}
\hline
Method & mIOU (\%) \\
\hline\hline
FCN-8s \cite{long2015fully} & 37.8 \\
CRF-RNN \cite{zheng2015conditional} & 39.3 \\
HO\_CRF \cite{arnab2016higher} & 41.3 \\
Piecewise \cite{lin2016efficient} & 43.3 \\
VeryDeep \cite{wu2016bridging} & 44.5 \\
DeepLabv2 \cite{chen2018deeplab} & 45.7 \\
RefineNet \cite{lin2017refinenet} & 47.3 \\
EncNet \cite{zhang2018context} & 51.7 \\
\hline
Our proposed (Xception-65) & 51.4 \\
Our proposed (Xception-71) & \textbf{52.5} \\
\hline
\end{tabular}
\end{center}
\caption{State-of-the-art methods on PASCAL Context {\it val} set.}
\label{table:state_of_the_art_methods_on_pascal_context}
\end{table}
Finally, we compare the framework of our proposed decoder with state-of-the-art methods. To compete with these state-of-the-art methods, we choose Xception-65 as the backbone network and the best feature aggregation in the ablation study for our decoder.

Following previous methods, SBD \cite{hariharan2011semantic} and COCO \cite{lin2014microsoft} are used to train the model as well. Specifically, the model is successively trained over COCO, SBD and PASCAL VOC {\it trainval} set, with the training protocol described in implementation details. Each round is initialized with the last round model and the base learning rate is reduced accordingly (i.e. 0.007 for COCO, 0.001 for SBD and 0.0001 for {\it trainval}). We use $500k$ iterations when training over COCO and $30k$ iterations for the last two rounds. Additionally, following previous works \cite{chen2017rethinking, chen2018encoder}, we make use of multi-scale testing and left-right flipping when inferring over {\it test} set.

As shown in Table \ref{table:state_of_the_art_methods}, our framework sets the new record on PASCAL VOC and improve the previous method DeepLabv3+ with the same backbone by 0.3\%, which is significant due to the benchmark has been very competitive. Meanwhile, since our proposed decoder can eliminate the need for high-resolution feature maps, we employ output stride being 16 instead of 8 in DeepLabv3+ when inferring over {\it test} set. As a result, our whole framework only takes 30\% computation of DeepLabv3+ (897.94B vs. 3055.35B in Multiply-Adds) to achieve the state-of-the-art performance. The performance of our proposed framework on PASCAL Context {\it val} set is shown in Table \ref{table:state_of_the_art_methods_on_pascal_context}. With Xception-71 as backbone, our framework sets the new state-of-the-art on this benchmark dataset without pre-training on COCO.

\section{Conclusion}
We have proposed a flexible and light-weight decoder scheme for semantic image segmentation. This novel decoder employs our proposed DUpsampling to produce the pixel-wise prediction, which eliminates the need for computationally inefficient high-resolution feature maps from the underlying CNNs and decouples the resolution of the fused low-level features and that of the final prediction. This decoupling expands the design space of feature aggregation of the decoder, allowing almost arbitrary features aggregation to be exploited to boost the segmentation performance as much as possible. Meanwhile, our proposed decoder avoids upsampling low-resolution high-level feature maps back to the spatial size of high-resolution low-level feature maps, reducing the computation of decoder remarkably. Experiments demonstrate that our proposed decoder has advantages of both effectiveness and efficiency over the vanilla decoder extensively used in previous semantic segmentation methods. Finally, the framework with the proposed decoder attains the state-of-the-art performance while requiring much less computation than previous state-of-the-art methods.

{\bf Acknowledgments}
The authors would like to thank Huawei Technologies
for the donation  of GPU cloud computing resources.

{\small
\bibliographystyle{ieee}
\bibliography{egbib}
}

%

%
%
%
%
%
%
%

%

%

%
%

%
\section*{
Supplementary Material:
}

In this supplementary material, we 1) provide our result on PASCAL VOC \cite{everingham2010pascal} {\it test} set without COCO \cite{lin2014microsoft} pre-training and 2) showcase more visualization results of our proposed method.

\section{PASCAL VOC without COCO Pre-training}
In this experiment, following previous works \cite{wu2016wider, zhao2017pyramid, zhang2018context} without COCO pre-training, we train our model on SBD \cite{hariharan2011semantic} and then fine-tune it on official {\it trainval} set.
We use the same training protocol as described in the main paper.
The multi-scale testing and left-right flipping are employed when our model is evaluated on {\it test} set. No any post-processing is used. The final performance is obtained by uploading our test results to the official test server.
As shown in Table \ref{table:state_of_the_art_methods_wo_coco}, our proposed framework surpasses previous published methods by a large margin.

\begin{table}[!h]

\begin{center}
\small
\begin{tabular}{ r |c}
\hline
Method & mIOU (\%) \\
\hline\hline
DPN \cite{liu2015semantic} & 74.1 \\
Piecewise \cite{lin2016efficient} & 75.3 \\
ResNet-38 \cite{wu2016wider} & 82.5 \\
PSPNet \cite{zhao2017pyramid} & 82.6 \\
DFN \cite{yu2018learning} & 82.7 \\
EncNet \cite{zhang2018context} & 82.9 \\
\hline
Our proposed (Xception-65) & \textbf{85.3} \\
\hline
\end{tabular}
\end{center}
\caption{State-of-the-art methods on PASCAL VOC {\it test} set without COCO pre-training.}
\label{table:state_of_the_art_methods_wo_coco}
\end{table}

\section{Visualization}
\begin{figure*}[!t]
  \centering
  \includegraphics[width=.8\linewidth]{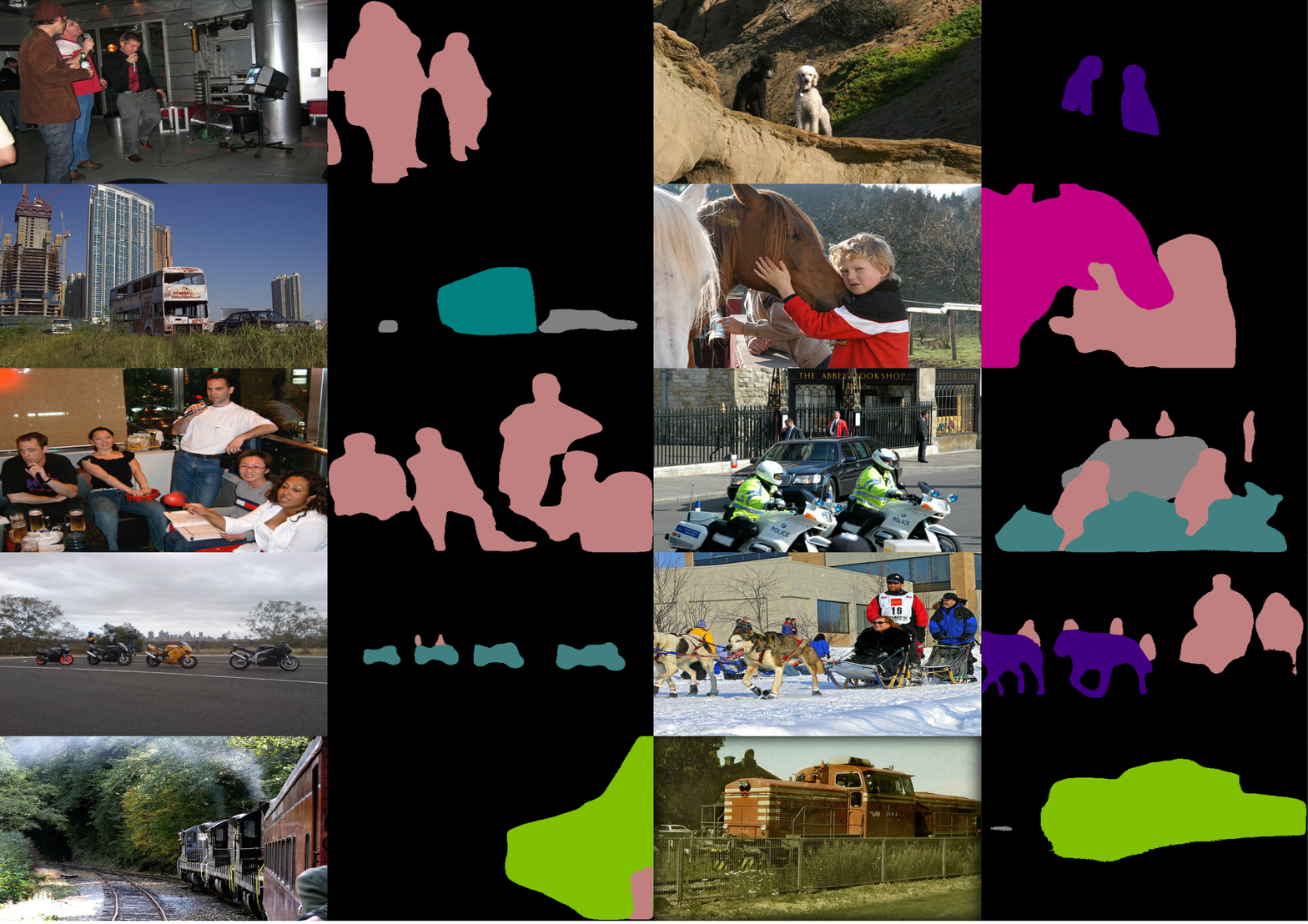}
  \caption{Visualization results from {\it val} set. The proposed method works reliably in a lot of challenging cases including small, distant and incomplete objects.}
  \label{fig:prediction_results1}
\end{figure*}
\begin{figure*}[!t]
  \centering
  \includegraphics[width=.8\linewidth]{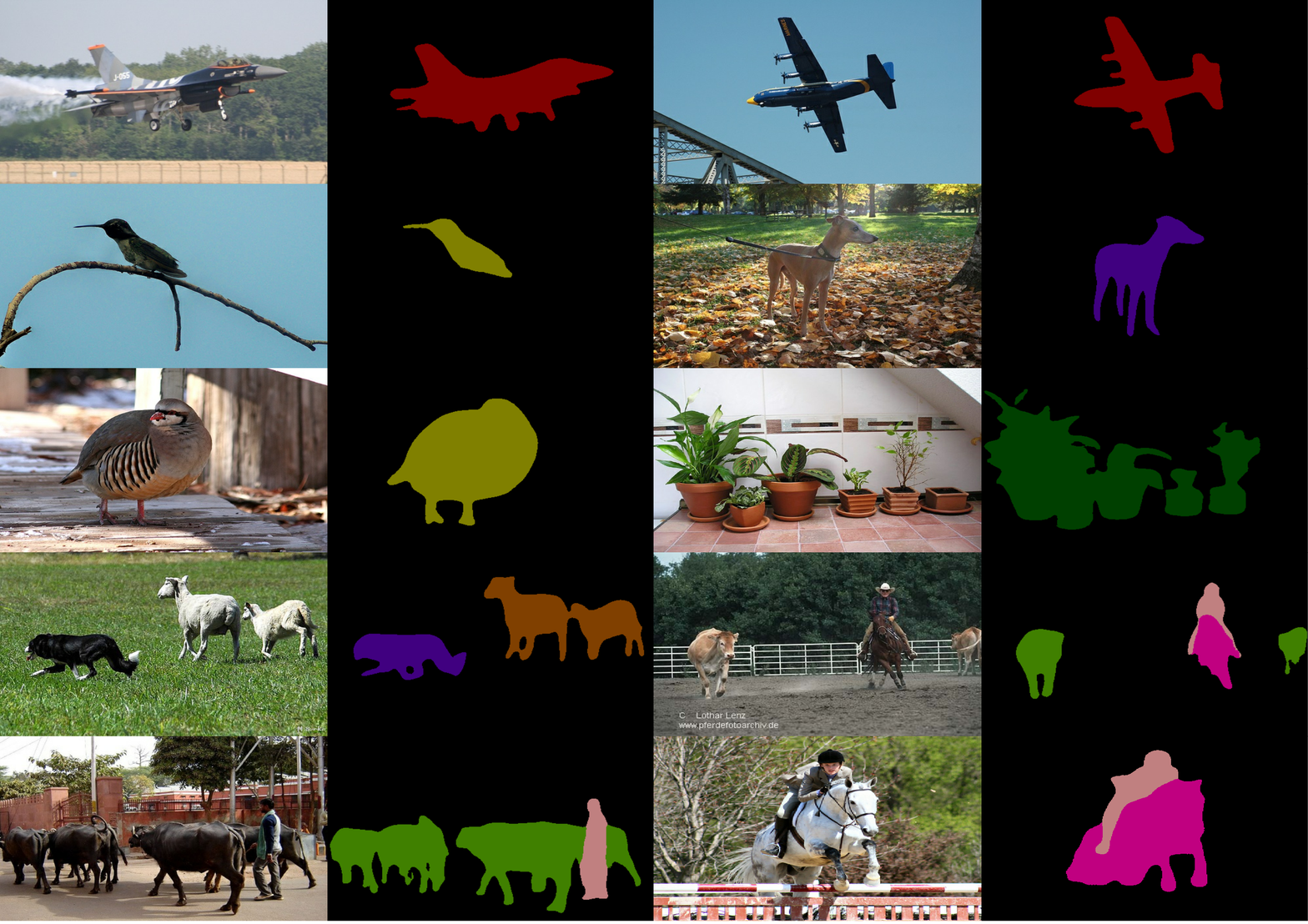}
  \caption{Visualization results from {\it val} set. The proposed method can yield fine-grained segmentation, with low-resolution CNNs output feature maps.}
  \label{fig:prediction_results2}
\end{figure*}
The visualization results of our method are shown in Fig. \ref{fig:prediction_results1} and Fig. \ref{fig:prediction_results2}.
As shown in Fig. \ref{fig:prediction_results1}, without any post-processing, the proposed method works very well in a lot of challenging cases. Small, distant and incomplete objects can be segmented well.
Meanwhile, as shown in Fig. \ref{fig:prediction_results2}, although we employ ``output stride" being 16 when evaluating, which results in low-resolution CNNs output feature maps, our model can still yield fine-grained segmentation due to the use of proposed DUpsamling.

\end{document}